\documentclass[journal,twoside,web]{ieeecolor}
\pdfoutput=1
\usepackage{generic}
\usepackage{cite}
\usepackage{amsmath,amssymb,amsfonts}
\usepackage{algorithmic}
\usepackage{graphicx}
\usepackage{algorithm,algorithmic}
\usepackage{hyperref}
\hypersetup{hidelinks=true}
\usepackage{booktabs}
\usepackage{multirow}
\usepackage{textcomp}
\usepackage{makecell}
\usepackage{graphicx}
\usepackage{textcomp}
\usepackage{array} 
\usepackage{longtable}
\usepackage{booktabs}
\usepackage{float}
\usepackage{amsmath}
\usepackage{lipsum}
\usepackage{graphicx}
\usepackage{bm}
\usepackage{siunitx}
\usepackage{makecell}

\usepackage[caption=false, font=footnotesize]{subfig}

\def\BibTeX{{\rm B\kern-.05em{\sc i\kern-.025em b}\kern-.08em
    T\kern-.1667em\lower.7ex\hbox{E}\kern-.125emX}}
\begin{document}
\title{Federated Learning for Medical Image Classification: A Comprehensive Benchmark}
\author{Zhekai Zhou, Guibo Luo, Mingzhi Chen, Zhenyu Weng, and Yuesheng Zhu
\thanks{Corresponding authors: Guibo Luo and Yuesheng Zhu.}
\thanks{Zhekai Zhou, Guibo Luo, Mingzhi Chen, and Yuesheng Zhu are with the School of Electronic and Computer Engineering, Peking University, Shenzhen, China, 518055 (e-mail: zhekai08@pku.edu.cn; luogb@pku.edu.cn; 2301212726@stu.pku.edu.cn; zhuys@pku.edu.cn). }
\thanks{Zhenyu Weng is with the Shien-Ming Wu School of Intelligent Engineering, South China University of Technology, Guangzhou, China, 510640 (e-mail: wzytumbler@gmail.com).}
}

\maketitle
\begin{abstract}
The federated learning paradigm is well-suited for the field of medical image analysis, as it can effectively cope with machine learning on isolated multi-center data while protecting the privacy of participating parties. However, current research on optimization algorithms in federated learning often focuses on limited datasets and scenarios, primarily centered around natural images, with insufficient comparative experiments in medical contexts. In this work, we conduct a comprehensive evaluation of several state-of-the-art federated learning algorithms in the context of medical imaging. 
We conduct a fair comparison of classification models trained using various federated learning algorithms across multiple medical imaging datasets. Additionally, we evaluate system performance metrics, such as communication cost and computational efficiency, while considering different federated learning architectures.
Our findings show that medical imaging datasets pose substantial challenges for current federated learning optimization algorithms. No single algorithm consistently delivers optimal performance across all medical federated learning scenarios, and many optimization algorithms may underperform when applied to these datasets.
Our experiments provide a benchmark and guidance for future research and application of federated learning in medical imaging contexts. Furthermore, we propose an efficient and robust method that combines generative techniques using denoising diffusion probabilistic models with label smoothing to augment datasets, widely enhancing the performance of federated learning on classification tasks across various medical imaging datasets.
Our code will be released on GitHub, offering a reliable and comprehensive benchmark for future federated learning studies in medical imaging.
\end{abstract}


\section{Introduction}
\label{sec:introduction}
\IEEEPARstart{T}{raditional} machine learning techniques utilize datasets that have been collected and organized to learn data features and develop models for tasks such as regression and classification. The broader the data sources and the larger the dataset, the better the generalizability and usability of the resulting models. When constructing health prediction models using medical data like electronic health records or medical imaging, training solely on data from a specific population may result in models that are not applicable to other groups~\cite{obermeyer2019dissecting}. However, the scope of medical datasets often exceeds the capacity of a single hospital or city, making centralized data pre-processing and model training challenging. Furthermore, collecting data for training models poses risks of privacy breaches. Although there is a desire to create unified and effective models for prediction, participants may be reluctant to share data due to privacy concerns, which could violate laws or regulations. Medical data, such as electronic health records and medical imaging, must be safeguarded to protect patient privacy. These issues contribute to the phenomenon of data silos. In response, Google first proposed the concept of federated learning (FL) in 2016\cite{mcmahan2017communication}, which combines distributed machine learning with privacy protection to address the data silo problem. The core idea is that participants maintain their original data locally without exchanging or sharing them, while collaboratively training by transmitting the machine learning model between participants. In the context of medical imaging, machine learning tasks often align with the application scenarios of horizontal federated learning~\cite{yang2019federated}, where participants share the same feature space but have different sample sets. For instance, hospitals in different cities may have test records that include data from different populations diagnosed with the same disease. Therefore, in this work, we focus on the application of horizontal federated learning in medical image classification tasks.

\textbf{Motivation of our study}: A key challenge in horizontal federated learning is the non-IID (non-independent and identically distributed) data among multiple participants. The locally trained models from each client, based on heterogeneous data distributions, struggle to achieve convergence during training and maintain good performance after aggregation. Numerous federated learning algorithms have been developed to reduce model performance degradation caused by heterogeneity. However, previous research generally faces several challenges: 
(1) Limitations of Dataset Domains: Given the original intention of federated learning to protect privacy, its ideal application scenarios should involve highly sensitive tasks such as the classification or segmentation of medical imaging data. In such medical scenarios, the sub-datasets of each participant are typically small, derived from patients in distinct geographical regions, and the image data may be collected from various models of medical equipment. Consequently, these sub-datasets may exhibit certain differences in feature distribution. However, most existing studies have been tested primarily on traditional low-resolution visual classification datasets like handwritten digits dataset MNIST~\cite{lecun1998gradient} or natural image dataset CIFAR-10~\cite{krizhevsky2009learning}, which not only lack experimental evaluations on larger-sized images but also fail to validate the effectiveness of their methods in the field of medical imaging, e.g., recently proposed FedFed~\cite{yang2024fedfed} and FLUTE~\cite{liu2024federated}. 
(2) Absence of Real Multi-Center Datasets with Feature Heterogeneity: In the simulation of non-IID data scenarios, most studies opt to use Dirichlet distributions to model scenarios where sample labels are non-IID. This approach neglects the utilization of real multi-center distributed datasets and lacks testing for feature heterogeneity across images from different clients, e.g., FEDDECORR~\cite{shi2022towards}. 
(3) Simplicity of Classification Neural Networks Used: Many studies employ relatively simple neural network architectures, such as multilayer perceptrons (MLPs) or convolutional neural networks (CNNs) with fewer layers, which may struggle to handle the complexities of medical image classification tasks. For example, the recently proposed FedAS~\cite{yang2024fedas} and FedPAC~\cite{xu2023personalized}. 
(4) Insufficient Comparative Experiments: Previous research on optimization algorithms related to aggregation methods or client updates has lacked comparisons with newer optimization techniques, e.g., knowledge distillation. As a result, there is a notable absence of benchmark that integrates various types of federated learning optimization algorithms for analysis.

To bridge the aforementioned gaps, our work includes: (1) conducting experiments with various federated learning algorithms on multiple medical image classification datasets corresponding to real-world medical diagnostic tasks to extensively evaluate their performance; (2) constructing real multi-center distributed client datasets for federated learning simulations by utilizing the same type of medical images collected from hospitals in different regions; (3) employing a more complex architecture, i.e., ResNet-50 network~\cite{he2016deep}, as the classification model for experiments; (4) incorporating a range of algorithms, including traditional federated learning optimization, personalized federated learning, and one-shot federated learning, to establish a benchmark for comprehensive analysis of their strengths and weaknesses in terms of classification performance and the efficiency of the federated learning system; (5) proposing an efficient and robust method to enhance the performance of federated learning models across multiple medical datasets.

\section{Related Works}
\subsection{Federated Learning}
Federated learning is an emerging machine learning paradigm that coordinates multiple distributed clients (e.g., smart IoT devices and smartphones) for collaborative model training via an aggregation server~\cite{mcmahan2017communication}. By circumventing the high cost and privacy risks associated with collecting raw sensitive data from numerous clients, federated learning has garnered significant attention from researchers in recent years. It has been applied to various machine learning tasks, including image classification~\cite{oh2021fedbabu}, object detection~\cite{liu2020fedvision}, text classification~\cite{zhu2020empirical}, graph classification~\cite{xie2021federated}, and tasks related to medical data~\cite{yang2021flop,dayan2021federated}.

The ideal approach to federated learning involves directly transmitting the gradient values computed from each client's backpropagation steps to the server, which then performs average calculations for gradient descent updates on the global model, achieving unbiased collaborative training across multiple clients~\cite{mcmahan2017communication}. However, this method incurs excessive communication cost, making practical implementation challenging. Consequently, the initial optimization strategy, FedAvg, proposes conducting multiple rounds of stochastic gradient descent training locally on clients before uploading model parameters to the server for weighted averaging, thus reducing the substantial data transfer between clients and the server~\cite{mcmahan2017communication,lin2018don}. Nevertheless, in this setup, the statistical heterogeneity of local data across clients can hinder the convergence of federated learning process and the performance of the resultant global model~\cite{li2020federated,karimireddy2020scaffold}. The optimization objective for federated learning is to achieve good model performance on generalized testing data through effective local training on client's local data and efficient aggregation of client models. To meet this objective, various theoretical analyses have been proposed for optimization~\cite{stich2018local,li2019convergence,kairouz2021advances}, most of which emphasize the limitation of the ``client drift,'' namely the quick overfitting of local optimizations toward local optima, negating the efficiency of server aggregations. Current improvement pathways mainly include three directions detailed below.

\subsubsection{Traditional Optimization Approach}
This class of algorithms still based on the paradigm of FedAvg, solving the problem caused by heterogeneous datasets across clients by modifying the update details of the clients' local training and server models. At the client side, (1) the computed gradient values can be adjusted during local training to prevent potential gradient divergence~\cite{karimireddy2020scaffold}; (2) model output can be corrected to reduce the local model's confidence in poorly performing categories~\cite{li2021fedrs}; (3) model parameters can be modified to limit excessive updates based on limited local training dataset, thereby mitigating negative impacts on the global model~\cite{chen2023elastic}; and (4) local loss functions can be reformed to incorporate new terms that restrict local models from deviating too far from the global model~\cite{li2020federated, gao2022feddc, shi2022towards}. At the server side, modifications to the model parameter aggregation process can control the magnitude of updates to minimize the deviations between the client training results on heterogeneous data and the global optimum~\cite{karimireddy2020scaffold}. These strategies can also be effectively combined~\cite{acar2021federated,chen2023elastic}.

\subsubsection{Personalized Federated Learning}
Personalized federated learning assumes that it is challenging for a single model to perform well across all clients due to statistical heterogeneity. Therefore, each client maintains a usable model without solely relying on one single globally aggregated model to achieve their objectives. On the server side, this can be achieved by altering the aggregation method; for instance, by performing global weighted aggregation updates only on the classification heads of the final layers of the model~\cite{collins2021exploiting}, or by excluding batch normalization layers from the global aggregation updates~\cite{li2021fedbn}. On the client side, knowledge distillation techniques can be introduced to smooth the updates between local and global models~\cite{li2019fedmd,wu2022communication,xie2024mh}. In this setup, the local model acts as a teacher to guide the global model in learning local knowledge, while the global model serves as a teacher for the local model to acquire global knowledge.
Personalized tasks can also be associated with various subfields of machine learning, leading to multiple proposed formulations. 
For example, each local dataset in a federated setting can be regarded as a task within a multi-task learning framework. Consequently, the multi-task regularization methods can be applied to the personalized federated learning algorithms, e.g., FedEM~\cite{marfoq2021federated} and Ditto~\cite{li2021ditto}.

\subsubsection{One-Shot Federated Learning}
To improve communication efficiency, one-shot federated learning has emerged as a new paradigm. In this approach, clients perform pre-training solely on their local datasets and then engage in a single communication with the server to transmit all local models or relevant information.
The server subsequently addresses the heterogeneity among client datasets through techniques such as model ensemble or knowledge distillation~\cite{guha2019one,zhou2020distilled}, resulting in a globally applicable model with improved performance. This approach significantly reduces communication cost in federated learning. However, to enhance model performance on global data, most existing one-shot federated learning methods require the use of auxiliary datasets, such as leveraging unlabeled public data on the server for model augmentation or knowledge transfer~\cite{lin2020ensemble,deng2023enhancing}. Consequently, in scenarios like federated learning in healthcare, where training data are often scarce, these methods may struggle to achieve satisfactory results. Some studies have improved federated learning frameworks based on diffusion models\cite{yang2023one,yang2024exploring,zhang2023federated}. For instance, FedCADO~\cite{yang2023one} enables each client to train a classifier using local data, guiding the diffusion model to synthesize pseudo samples based on labels and classifier outputs for the training of global classification model. However, these studies have only explored the use of pre-trained stable diffusion models\cite{rombach2022high}. And they validated their effectiveness exclusively in the domain of natural images, without extending their application to federated learning in medical imaging.

Most studies on these federated learning optimization algorithms have not utilized real multi-center datasets, instead primarily validating their effectiveness on a limited range of natural image datasets.
Table~\ref{table1} presents the datasets utilized in experiments of recent optimization research on federated learning algorithms, with the bold entries indicating non-IID datasets with authentic heterogeneous distributions.

\renewcommand\arraystretch{1.5}%保证每列高度是原先的1.5倍
\begin{table*}%星号表示双栏
\caption{Datasets in Latest Federated Learning Algorithms Research}
\label{table1}
\centering
 \begin{tabular}{p{1.5cm}<{\centering}p{2cm}<{\centering}p{4cm}<{\centering}p{3cm}<{\centering}p{2cm}<{\centering}p{2.2cm}<{\centering}}%设定每列的宽度以及对齐方式，并且可以做到自动换行
 
  \Xhline{1.2pt}%第一条粗线
   
        & \textbf{FL Algorithms} & \textbf{Natural Image Datasets} & \textbf{Text Datasets} & \textbf{Recommendation System Datasets} & \textbf{Medical Imaging Datasets} \\ 
  \Xhline{1.2pt}%第二条粗线
    \multirow{9}*{\shortstack{Traditional\\Optimization\\Strategy}}
        & FedAvg & MNIST, CIFAR-10 & Shakespeare & -- & -- \\ 
        \cline{2-6}
        ~ & FedProx & MNIST, \textbf{FEMNIST} & Shakespeare, Sent140 & -- & -- \\
        \cline{2-6}
        ~ & FedDC & MNIST, CIFAR-10, CIFAR-100, EMNIST, Tiny-ImageNet & -- & -- & -- \\
        \cline{2-6}
        ~ & FedDERORR & CIFAR-10, CIFAR-100, Tiny-ImageNet & -- & -- & -- \\
        \cline{2-6}
        ~ & FedNova & CIFAR-10, Synthetic & -- & -- & -- \\
        \cline{2-6}
        ~ & SCAFFOLD & EMNIST & -- & -- & --  \\
        \cline{2-6}
        ~ & FedRS & MNIST, \textbf{FEMNIST}, CIFAR-10, CIFAR-100 & Shakespeare & MovieLen, Pinterest & -- \\
        \cline{2-6}
        ~ & FedDyn & MNIST, CIFAR-10, CIFAR-100 & -- & -- & -- \\
        \cline{2-6}
        ~ & Elastic Aggregation & MNIST, \textbf{FEMNIST}, CIFAR-10, CIFAR-100, CINIC-10 & Shakespeare & -- & -- \\
        \hline

    \multirow{4}*{\shortstack{Personalized \\ FL}}
        & FedBN & \textbf{Office-Caltech10}, \textbf{DomainNet} & -- & -- & ABIDE I \\
        \cline{2-6}
        ~ & FedMD & MNIST, \textbf{FEMNIST}, CIFAR-10, CIFAR-100 & -- & -- & -- \\
        \cline{2-6}
        ~ & FedKD & -- & ADR & MIND & -- \\
        \cline{2-6}
        ~ & Ditto & Fashion MNIST, \textbf{FEMNIST} & -- & -- & -- \\
        \hline

    \multirow{3}*{\shortstack{One-Shot \\ FL}}
        & OSFL & EMNIST & Sent140 & -- & -- \\
        \cline{2-6}
        ~ & FedDF & CIFAR-10, CIFAR-100, ImageNet & AG News, SST2 & -- & -- \\
        \cline{2-6}
        ~ & DOSFL & MNIST & IMDB, TREC-6, Sent140 & -- & -- \\
  \Xhline{1.2pt}%第三条粗线
  \end{tabular}
\vspace{-0.8em}
\end{table*}

\subsection{Federated Learning Benchmark}
Currently, there are efforts on benchmark in federated learning concerning heterogeneous datasets, e.g., LEAF~\cite{caldas2018leaf}, which includes human face image dataset CelebA~\cite{liu2015faceattributes} and natural image datasets like CIFAR-10~\cite{krizhevsky2009learning}, as well as text datasets like Sentiment140~\cite{go2009twitter}. Additionally, there are benchmark studies focused on heterogeneous systems, such as Flower~\cite{beutel2020flower} and FedML~\cite{he2020fedml}. Specific benchmarks for federated learning in specialized applications have also emerged, including FS-G~\cite{wang2022federatedscope} for graph neural network federated learning and pFL-Bench~\cite{chen2022pfl} for personalized federated learning, as shown in Table~\ref{table2}. However, there is still a lack of comprehensive federated learning benchmark focusing on applications in the medical imaging field.

\begin{table}
\caption{Datasets in Existing Federated Learning Benchmarks}
\label{table2}
\centering
\setlength{\tabcolsep}{3pt}
\begin{tabular}{|p{40pt}|p{75pt}|p{50pt}|p{50pt}|}
\hline
Benchmarks & Natural Image Datasets & Text Datasets & Graph Datasets \\
\hline
LEAF & FEMNIST, CelebA & Sent140, Shakespeare, Reddit & -- \\
\hline
pFL-Bench & FEMNIST, CIFAR-10 & COLA, SST-2, Twitter & Cora, CiteSeer, CiteSeer \\
\hline
Flower & FEMNIST, CIFAR-10 & Shakespeare & -- \\
\hline
FedML & MNIST, FEMNIST, CIFAR-100 & Shakespeare, StackOverflow & -- \\
\hline
FS-G & -- & -- & Cora, CiteSeer, CiteSeer \\
\hline
\end{tabular}
\vspace{-0.8em}
\end{table}

\subsection{Federated Learning on Medical Imaging Data}
Early work has applied the foundational FedAvg paradigm to medical imaging datasets, analyzing its effectiveness through experiments, e.g., breast density classification on the BI-RADS dataset~\cite{roth2020federated}. Currently, there are also some researches trying to make some optimization based on the characteristics of the medical domain. For example, CusFL~\cite{wicaksana2022customized} conducted experiments on the HAM10K~\cite{tschandl2018ham10000} dataset utilizing dermoscopic images for melanoma diagnosis, and FedNI~\cite{peng2022fedni} conducted experiments using the Brain fMRI dataset ABIDE~\cite{di2014autism} to verify the effectiveness on ASD classification tasks. Dermoscopic image dataset MSK~\cite{codella2018skin} is also involved in federated learning experiments~\cite{chen2022personalized}. Additionally, some studies have explored the application of federated learning in COVID-19 diagnosis, primarily utilizing chest X-ray images from the COVID-FL~\cite{yan2023label} or COVIDx~\cite{wang2020covid,yang2021flop} datasets, as well as in diabetic classification tasks using color retinal images from the Retina dataset~\cite{yan2023label,diabetic-retinopathy-detection}, and for breast tumor pathology image classification with the BreakHis dataset~\cite{spanhol2016breast, agbley2023federated}. Some of the used medical datasets are shown in Table~\ref{table3}. However, there remains a lack of benchmark that comprehensively evaluate multiple types of federated learning optimization methods across medical image classification tasks in multiple disease diagnoses fields.

\begin{table}
\caption{Datasets in Existing Research on Federated Learning for Medical Image Analysis}
\label{table3}
\centering
\setlength{\tabcolsep}{3pt}
\begin{tabular}{|p{35pt}|p{42pt}|p{70pt}|p{50pt}|}
\hline
Dataset & Modality & Purpose & Resolution \\
\hline
COVIDx & Chest X-Ray & COVID-19 diagnosis & $> 1000 \times 960$ \\
\hline
HAM10K & Dermoscopy & Skin lesion classification & $600 \times 450$ \\
\hline
Retina & Color retinal image &  Diabetic Retinopathy detection & $> 3000 \times 2000$ \\
\hline
ABIDE & Brain fMRI & ASD classification & -- \\
\hline
BreakHis & Microscopic image of pathological tissue & Breast tumor classification & $700 \times 460$ \\
\hline
\end{tabular}
\vspace{-0.8em}
\end{table}

\section{Federated Learning Benchmark Setting}
\subsection{Generalized Federated Learning Setting}
Federated learning is defined as follows: there are $K$ participating clients, each possessing a local dataset $D_k$ consisting of $n_k$ samples. Each client trains a local model $W_k$, and the server aggregates these to obtain a global model $W$. The objective is for $W$ to achieve optimal performance when tested on the collective dataset formed by all $D_k$. Given a loss function $L$, the classical formulation of the neural network optimization problem in a federated learning setup can be defined as
\begin{equation}W^{\ast} = \mathop{\arg\min}\limits_{w \in \mathcal{W}}\sum\limits_{k=1}^{K}\sum\limits_{i=1}^{n_k}L(w, x_{k,i}, y_{k,i}).\label{eq1}\end{equation}

\textbf{Federated Averaging (FedAvg)}~\cite{mcmahan2017communication} is the most commonly used baseline algorithm in horizontal federated learning. FedAvg proposes that multiple participating clients perform several rounds of stochastic gradient descent training on their local datasets. 
A central server then collects the model parameters from each client and computes the global model parameters through a weighted average, where the weights correspond to the size of each client's dataset, as shown in (\ref{eq2}), where $p_k$ is an aggregation weight and $t$ denotes the global round of federated learning. The global model is subsequently transmitted back to the clients for further local training.
\begin{equation}\label{eq2}
    W^{t+1}=\sum\limits_{k=1}^{K}p_k \cdot W_k^t = \sum\limits_{k=1}^{K} \frac{n_k}{\sum\limits_{k=1}^{K} n_k} \cdot W_k^t
\end{equation}
 
The advantage of FedAvg is that it reduces the number of communication rounds required to achieve model convergence by allowing participants to conduct multiple rounds of local training before sharing their model parameters. However, a potential problem is that the weighted average of the locally optimal model parameters may deviate from the optimal point on the overall dataset, which is not the ideal objective of federated learning. This problem is particularly acute in non-IID data scenarios. Various federated learning algorithms have been developed in recent years to address this deficiency.

\subsection{Comparison of Existing Optimzation Algorithms}
Our benchmark work first includes five traditional federated learning optimization algorithms.

\textbf{FedProx}~\cite{li2020federated} improves upon FedAvg by refining the clients' local optimization objectives, directly constraining the magnitude of local model updates. It introduces an additional $L2$ regularization term in the local loss function to limit the distance between the local model and the global model, thereby avoid excessive local updates causing significant deviations from the global optimum during aggregation. The expression for the client-side local loss function is shown in (\ref{eq4}), where $\mu$ is a hyperparameter to control the weight of the added term in loss function. Overall, the modifications made by FedProx are lightweight and easily implementable in practice. A notable drawback is that users may need to carefully tune the value of $\mu$ to achieve satisfactory accuracy. If $\mu$ is set too low, the regularization term has little effect; conversely, if $\mu$ is too high, local updates become very small, resulting in slow convergence.
\begin{equation}\label{eq4}
    \mathcal{L}_k(w; w_k^t) = L(w, x_{k,i}, y_{k,i}) + \mu \Vert w - w_k^t \Vert ^2
\end{equation}

\textbf{MOON}~\cite{li2021model} leverages the principles of contrastive learning. It introduces a novel loss term known as model-contrastive loss, which calculates the similarity between three representations during the local training phase, as shown in (\ref{eq5}). This formulation enables the current local model to achieve better performance during iterative local training in two ways. First, it encourages the representations produced by the local model to become closer to those of the previous global model. Second, it drives the representations produced by the local model away from those of the previous round local model. A hyperparameter $\mu$ is introduced to control the balance between the new loss term and the cross-entropy loss term.
\begin{equation}\label{eq5}\begin{aligned}
    \mathcal{L} = & L(w, x_{k,i}, y_{k,i}) - \\ & \mu \log\frac{\exp({\rm sim}(z, z_{\textit{glob}})/\tau)}{\exp({\rm sim}(z, z_{\textit{glob})}/\tau) + \exp({\rm sim}(z, z_\textit{prev})/\tau)}
\end{aligned}\end{equation}

\textbf{FedNova}~\cite{wang2020tackling} proposes that, due to differences in local datasets, some clients may produce larger local model updates during training. If these updates are simply averaged to update the global model, they can disproportionately affect the global parameters. To ensure unbiased global updates, FedNova normalizes each client's local updates based on the number of local steps taken before aggregating them, and scales them according to hyperparameters, implicitly modifying the aggregation weights among clients.

\textbf{FedRS}~\cite{li2021fedrs} proposes that when data across different clients is in a heterogeneous state, the \textit{Softmax} layers in each local model can lead to poor classification performance in the average aggregated global model. To mitigate this problem, FedRS introduces a constrained \textit{Softmax} approach, which adds weight parameters to the \textit{Softmax} function to restrict updates in directions that would result in inaccurate classifications.

\textbf{Elastic Aggregation}~\cite{chen2023elastic} introduces the concept that heterogeneity in client data leads to discrepancies in local model parameters and oscillations during average aggregation. Therefore, the parameters that are less sensitive to the variation of model output can be updated more freely to minimize the local loss of a single client, while minimizing updates to more sensitive parameters. It employs the $L2$ norm of the gradient value to assess the sensitivity of the objective function output to variations in network parameters. At the beginning of each local training round, this sensitivity is calculated based on the current model on a subset of the local dataset. A control coefficient is derived from the sensitivity of the parameters of each network layer, which is then used during server aggregation to scale down the updates to the global model's parameters for that layer, thereby reducing the oscillations that can result from straightforward averaging.

Then, the following two are personalized federated learning algorithms involved by our benchmark experiments.

\textbf{FedBN}~\cite{li2021fedbn} is a classic personalized federated learning algorithm emphasizes the feature shift problem, which is a critical challenge in applying federated learning in many real-world scenarios in medical imaging. For instance, in a diagnosis task, medical images like MRI images collected from various hospitals may exhibit uniformly distributed labels. However, due to the differences in imaging machines used by hospitals, the appearance of the images, or specifically the features, can vary significantly. This necessitates the resolution of adverse effects on the global model aggregation in such feature-heterogeneous scenarios. FedBN highlights that the application of batch normalization layers in the client models can effectively mitigate the impact of feature shift, and it proposes that when local models utilize batch normalization method, the parameters of these layers (e.g., running mean and variance) should be excluded from the aggregation process. The modifications introduced by FedBN are lightweight, as they only alter the scope of model parameter aggregation.

\textbf{Personalized Retrogress-Resilient FL}~\cite{chen2022personalized} introduces a personalized federated learning architecture based on knowledge distillation. It is noted that directly updating the local models with the aggregated server parameters can lead to a loss of learned local knowledge, further diminishing the performance of the personalized model in subsequent local iterations. Therefore, the algorithm proposes to smoothly integrate global knowledge with local priors via Deputy-Enhanced Transfer, rather than replacing them outright. Each client includes a deputy that receives the aggregated parameters from the server. The whole local process then involves three steps: Recover, Exchange, and Sublimate, which facilitate the gradual transfer of global knowledge from the deputy to the personalized local model by knowledge distillation. The delineation of different phases is based on the performance of the current deputy and the personalized model on the local training set. In each phase, the loss function is composed of distinct components, including the cross-entropy term and the KL divergence term. 

Finally, our benchmark involves a one-shot federated learning algorithm. \textbf{DENSE}~\cite{zhang2022dense} introduces a data-free one-shot federated learning framework, distinguishing itself from traditional federated learning approaches that involve multiple global rounds. DENSE performs model aggregation only once, significantly reducing the communication cost associated with transmitting model parameters between clients and the server. In its framework, each client first conducts pre-training on its local data. Subsequently, the server trains the global model through two phases: a data generation phase and a model distillation phase. In the first phase, an ensemble of local models uploaded by clients is used to train a generator, which creates synthetic data for use in the second phase. During the second phase, the generated pseudo-samples are employed for distillation learning, allowing the knowledge from the ensemble model to be transferred into the global model.

\subsection{Proposed Method}
\begin{figure}[!t]
\centerline{\includegraphics[width=\columnwidth]{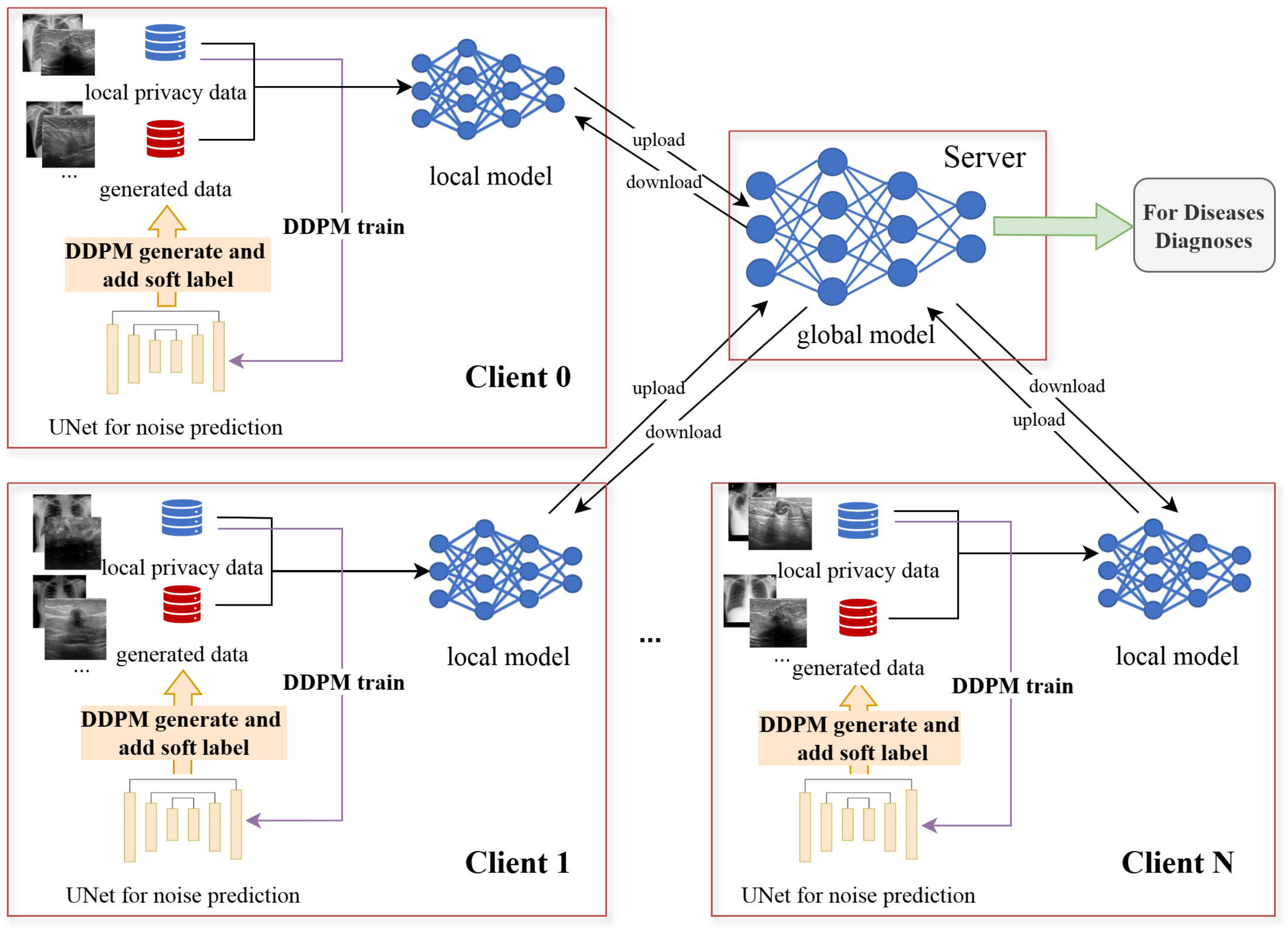}}
\caption{Architecture of our proposed federated learning optimization method. Our method incorporates the generative technique of DDPM and label smoothing within the FedAvg framework.}
\label{fig0}
\vspace{-0.8em}
\end{figure}

To further improve the performance of federated learning in medical image analysis and to ensure its wide applicability in the field of medical imaging, we introduce a novel and concise idea of federated learning optimization in our benchmark, as shown in Fig.~\ref{fig0}. In many federated learning scenarios, the size of client local dataset is relatively small, which may lead to insufficient training and hinder improvements in accuracy. Therefore, we propose to train an image generation model using the local data and leverage the trained model to generate additional images to supplement the federated learning training set. We employ the classic conditional Denoising Diffusion Probabilistic Models (DDPM)~\cite{ho2020denoising} for this purpose. By utilizing the unique image datasets from different clients, we train stable and high-quality generative models from scratch, thereby achieving personalized image generation. Moreover, since each client trains the DDPM exclusively on local data and the generated data remains on the local device, client data privacy is thoroughly protected, meeting the requirements of the federated learning setting.

\subsubsection{DDPM}
The objective of the Conditional DDPM is to generate target images under given specific conditions, which in our application correspond to the category labels of the images. The diffusion model comprises two processes: a forward noise-adding process and a reverse denoising process. In the forward process, Gaussian noise is progressively introduced to the images in the training set over several iterations, causing them to gradually lose their distinguishable features until they converge to a standard normal distribution. This noise addition is a Markov process that incrementally alters the original image. At each time step $t$, standard Gaussian noise is added to the previous round of noisy resultant image, guided by a parameter $\beta_t$ that grows linearly with the time step, as in \begin{equation} \mathbf{x}_t = \sqrt{1 - \beta_t} \cdot \mathbf{x}_{t-1} + \sqrt{\beta_t} \cdot \mathbf{\epsilon}, \mathbf{\epsilon} \sim \mathcal{N}(\mathbf{0}, \mathbf{I}).\label{eq6}\end{equation} The goal of the reverse process is to start with a noise image drawn from a standard normal distribution and to fit the noise distribution added at each step of the forward process in order to remove it, effectively reversing the noise addition process. Ultimately, this results in an approximation of the original image distribution. The reverse process employs a neural network to predict the noise added at each step based on the resultant image generated of each step in the forward process. It allows for the removal of the predicted noise distribution, transforming a noise image that conforms to a standard Gaussian distribution into the target image step by step. Given the similarity between the DDPM task and image denoising, we have chosen U-Net~\cite{ronneberger2015u} as the model architecture for noise prediction during training. U-Net features a U-shaped network structure composed of interconnected encoder and decoder layers. The encoder down-samples the original image to obtain a feature representation, while the decoder up-samples this feature to predict the target noise distribution.

\subsubsection{Label Smoothing}
In practical scenarios, limited training data and the uncertainty of the DDPM training process from scratch may lead to discrepancies between the data distribution generated by models and the real data distribution. Such discrepancies can mislead the classification model into learning incorrect information, ultimately resulting in a decline in performance. For instance, model collapse is observed when iteratively trained with generated data~\cite{wang2024generated}. Similarly, directly incorporating generated data into the training set led to a decrease in classification performance on CIFAR-10 in some experiments\cite{shumailov2024ai}.
To mitigate the negative impact of discrepancies between the distribution of generated data and the original data distribution on the training of the classification models, we employ the label smoothing technique\cite{muller2019does,szegedy2016rethinking} to reconstruct the labels of the generated data. It transforms the original one-hot encoded labels into probability distribution vectors for each category, thereby increasing the entropy of the generated data labels. The formula representation of soft label is shown as (\ref{eq7}), where $K$ denotes the number of classification categories in the dataset, $\mathbf{y}$ denotes the original one-hot encoded label, and $\alpha \in (0, 1)$ is the smoothing parameter. This approach prevents the model from becoming overly confident in assigning a sample to a specific category during the federated learning process, thereby reducing the risk of overfitting on synthetic data and improving the generalization ability of the client models on the real global dataset.
\begin{equation}\label{eq7}
    \mathbf{y}^{soft} = (1-\alpha) \cdot \mathbf{y} + \frac{\alpha}{K} \cdot \bm{1}
\end{equation}

\section{Experiments}

\subsection{Datasets}
\subsubsection{Introduction of Used Datasets}
We selected several medical imaging datasets collected from hospitals for our experiments, aiming to simulate federated learning applications in real-world medical scenarios.

\textbf{ColonPath}~\cite{wang2023real}: Lesion tissue screening in pathology patches. The dataset consists of a total of 10,009 large tissue patches images obtained from endoscopic pathology examinations, with an average size of $1024 \times 1024$. Based on existing lesion area labels, the data are categorized into positive and negative classes, corresponding to tissue with and without lesions, respectively.

\textbf{NeoJaundice}~\cite{wang2023real}: Neonatal jaundice evaluation in skin photos. The dataset comprises a total of 2,235 photographs, with an average size of $567 \times 567$. Recent image classification techniques are utilized to estimate the levels of total serum bilirubin from images of infants' heads, faces, and chests. The samples are categorized into low and high total serum bilirubin based on the corresponding actual test results.

\textbf{Retino}~\cite{wang2023real}: Diabetic Retinopathy (DR) grading in retina images. The dataset comprises a collection of retinal images that can be used for the grading and assessment of DR, which can lead to vision loss and blindness in diabetic patients. This dataset is utilized to train machine learning models for the automatic detection and diagnosis of the severity of DR, enabling early identification of lesions. The dataset contains a total of 1,392 images, with an average size of $2736 \times 1824$. The samples are categorized into five classes: No DR, Mild, Moderate, Severe, and Proliferative DR.

\textbf{DR}~\cite{aptos2019-blindness-detection}: Similar to Retino, the APTOS 2019 Blindness Detection Dataset comprises a collection of fundus images collected over an extended period under various conditions and environments, primarily aimed at diagnosing Diabetic Retinopathy. The dataset includes a total of 3,662 samples, divided into a training set of 2,930 images, a validation set of 366 images, and a test set of 366 images. The samples are also categorized into five classes based on the severity of Diabetic Retinopathy as those in Retino.

\textbf{CRC}~\cite{kather2018100}: Human colorectal cancer detection from image patches. The training set NCT-CRC-HE-100K comprises a collection of 100,000 non-overlapping tissue image patches extracted from 86 human cancer tissue slides and normal tissues stained with hematoxylin and eosin (H\&E). The test set CRC-VAL-HE-7K includes 7,180 images derived from 50 colorectal adenocarcinoma patients, ensuring that the data is distinct from that of the training set. The samples are labeled into nine categories: adipose, background, debris, lymphocytes, mucus, smooth muscle, normal colon mucosa, cancer-associated stroma, and colorectal adenocarcinoma epithelium. This dataset is utilized for machine learning-based image classification, which can aid in the early detection and screening of colorectal cancer.

\textbf{COVID-QU-Ex}~\cite{tahir2021covid}: COVID-19 diagnosis from chest X-ray images. The dataset comprises a total of 3,728 chest X-ray images categorized into three classes: COVID-19, Non-COVID infections, and Normal, representing images of pneumonia caused by COVID-19, pneumonia due to other infections, and normal lung images, respectively. This dataset is employed for machine learning-based image classification, facilitating pneumonia diagnosis through chest X-ray analysis.

\textbf{Breast}~\cite{al2020dataset}: The Breast Ultrasound Images Dataset consists of a collection of breast ultrasound images utilized for early detection and identification of breast cancer through machine learning classification. The dataset includes 780 breast ultrasound images from 600 women aged between 25 and 75, with an average image size of $500 \times 500$. The images are categorized into three classes: Normal, Benign, and Malignant.

\begin{figure}[!t]
\centerline{\includegraphics[width=\columnwidth]{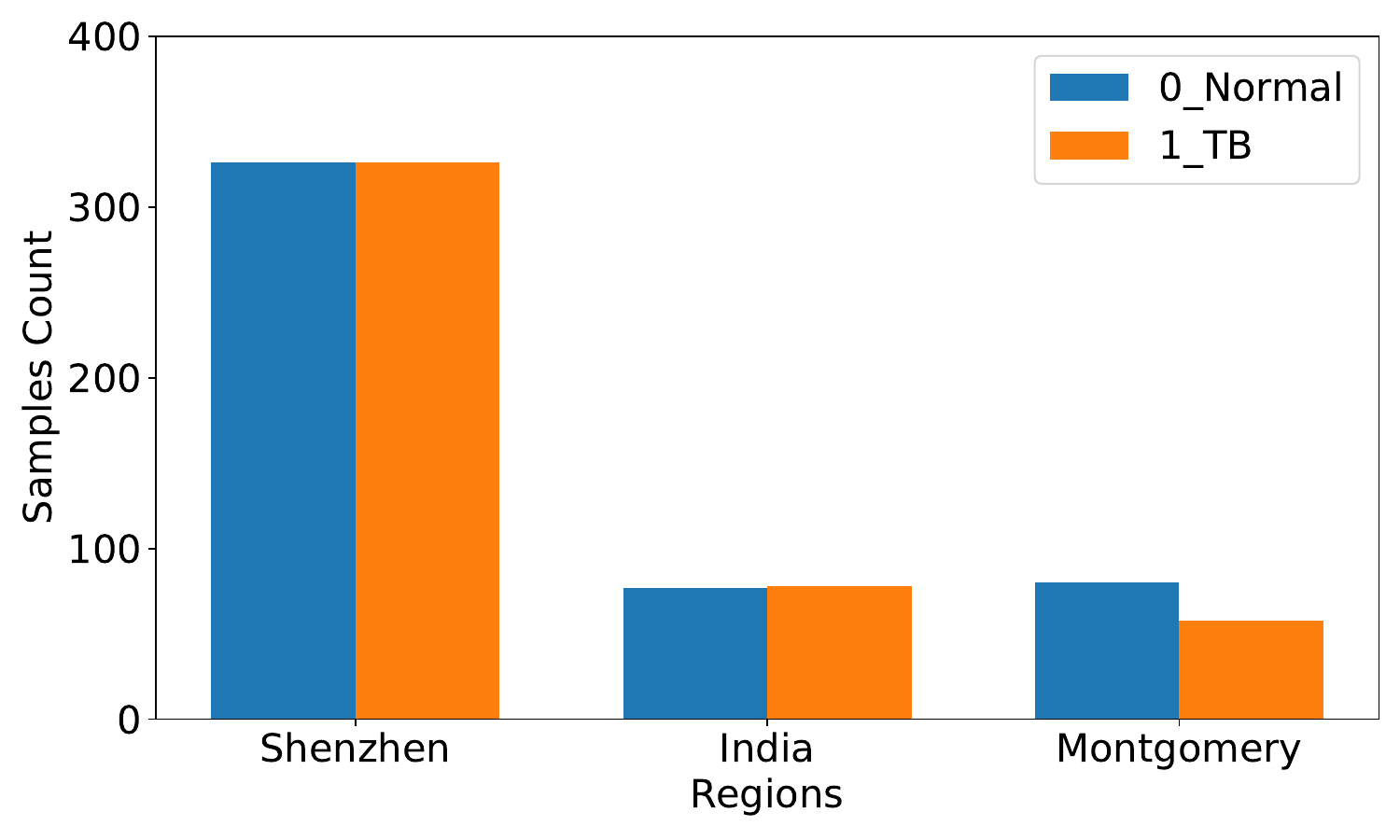}}
\caption{The sub-datasets from the three regions contain varying numbers of images, with an average samples count of 176 and a standard deviation of 222.3.}
\label{fig1}
\end{figure}

\begin{figure}[!t]
\centerline{\includegraphics[width=\columnwidth]{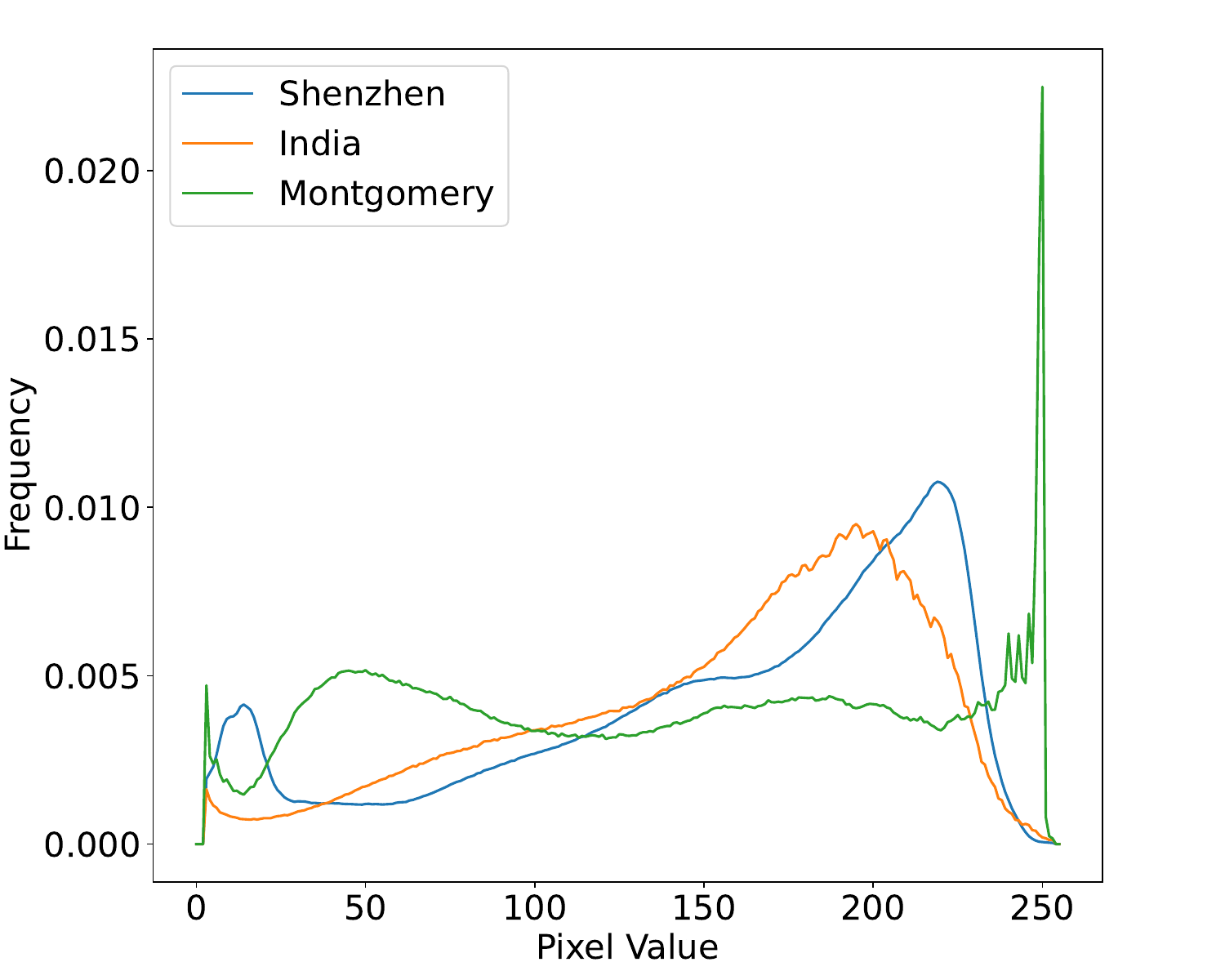}}
\caption{The distribution of multi-center datasets. The x-axis represents the pixel values, while the y-axis indicates their corresponding frequency across all images in the dataset. The curve reflects the distribution characteristics of the image pixels in the sub-dataset.}
\label{fig2}
\vspace{-0.8em}
\end{figure}

Furthermore, a real-world multi-center medical imaging dataset \textbf{TB}~\cite{rahman2020reliable} is utilized in our benchmark. To evaluate the performance of federated learning algorithms on real non-IID datasets, we collected three tuberculosis datasets from different regions: the Shenzhen dataset, the India dataset, and the Montgomery dataset. The Shenzhen dataset includes 471 training samples and 120 testing samples, the India dataset comprises 107 training samples and 34 testing samples, and the Montgomery dataset contains 90 training samples and 39 testing samples. The volume distribution of client data is presented in Fig.\ref{fig1}. Since they were collected from different cities and hospitals, the appearance and pixel distribution of the images varied considerably from one client to another, thus leading to the dissimilarity of image features across clients, namely feature shift. The pixel distribution of the client dataset is illustrated in Fig.\ref{fig2}.

\subsubsection{Data Pre-process}
\begin{figure*}[!t]
    \centering
    \subfloat[Original Number of Training Samples of Different Clients]{\label{subfigure5}\includegraphics[width=\columnwidth]{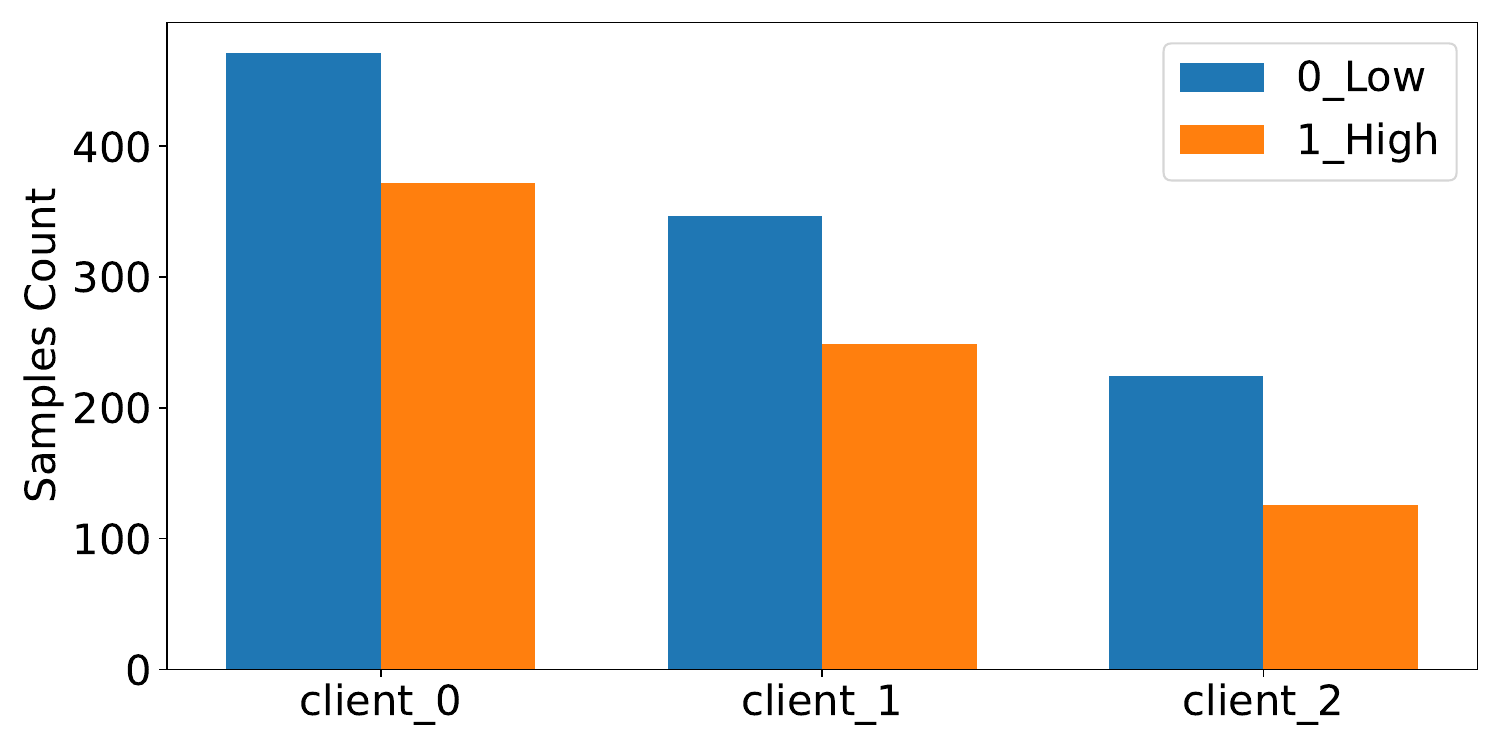}}
    \hfill
    \subfloat[Number of Training Samples of Different Clients After Incorporating Generated Data]{\label{subfigure6}\includegraphics[width=\columnwidth]{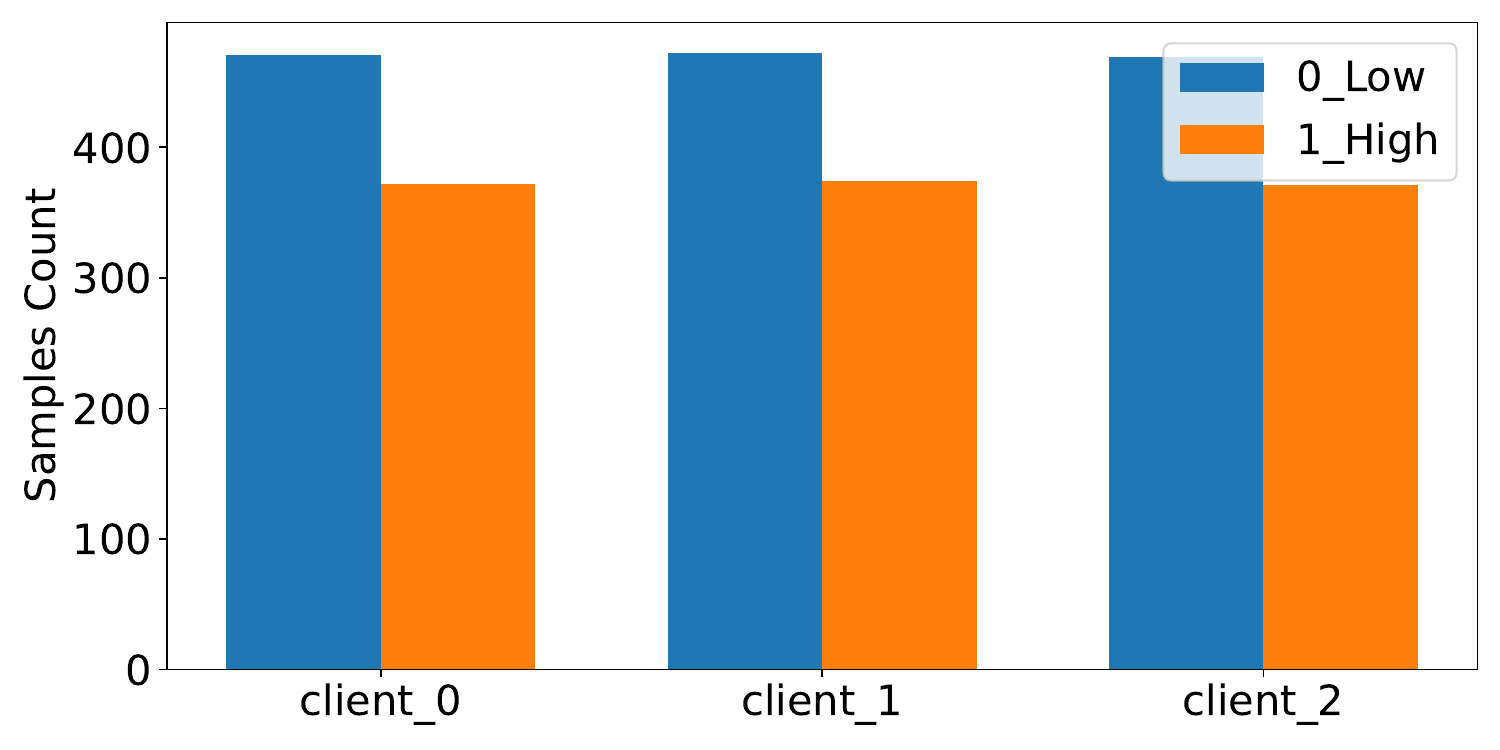}}
    
    \caption{The three partitioned client datasets from imbalanced NeoJaundice contain varying numbers of training samples, with an average of 596.3 samples and a standard deviation of 201.3. After utilizing DDPM for image generation, the training samples across clients are evenly distributed.}
    \label{fig3}
    \vspace{-0.8em}
\end{figure*}

For the datasets ColonPath, Retino, and NeoJaundice, which do not have predefined splits, we randomly divided the data into six-folds for cross-validation, with each having the same number of images. Five of the six folds are used as five clients' local training data while the remaining fold is used as the test set. The final performance of federated learning is determined by calculating the mean of the results obtained from the test sets of these six experiments. 
We also constructed a simulation reflecting the scenario of imbalanced local training data quantities among participants in real-world federated learning within the healthcare domain. The NeoJaundice dataset was unevenly partitioned across three clients, with the data distribution shown in Fig.\ref{subfigure5}. We transformed the size of images to $256 \times 256$ uniformly. And the pixel values are normalized between -1 and 1.

\subsection{Experiments Setup}

\subsubsection{Model Configuration}
A ResNet-50 model is employed for the classification of medical images, comprising four stages that contain 3, 4, 6, and 3 bottleneck blocks, respectively, and each bottleneck consists of 3 convolutional layers. The local model is initialized with pre-trained weights from ImageNet at the beginning of federated training. To accommodate the varying number of categories across different datasets, the dimension of final fully connected layer output is modified to correspond with the number of classes in the respective dataset. A challenge in implementing ResNet-50 within a federated learning framework is the aggregation of batch normalization layers. While local batch normalization layers can effectively handle local distributions, simple averaging of these parameters may not adequately capture the statistics of the global distribution, thereby introducing more instability. For instance, parameters named with ``num\_batches\_tracked'' should be excluded from the scope of aggregated parameters.

The training process of DDPM involves utilization of exponential moving average (EMA)~\cite{athiwaratkun2018there} to enhance the robustness of model. The U-Net includes three down-sample layers, three double convolution layers, and three up-sample layers, with the incorporation of self-attention mechanisms throughout. Each down-sampling layer doubles the number of channels and each up-sampling layer halves the number of channels. The entire network is composed of around 93M parameters.

\subsubsection{Hyperparameters}
Each participating client utilized an Adam optimizer with an initial learning rate of $\num{e-3}$, setting hyperparameters $\beta_1 = 0.9$, $\beta_2 = 0.999$, and weight decay of $\num{5e-4}$. The batch size for mini-batch gradient descent was set to 64, and each client was assigned a local training period of 5 epochs. Based on separate experiments and references to recommended values from the original studies, we set the hyperparameters involved in the optimization algorithms as detailed in Table\ref{table4}.
For DENSE framework, we pre-trained each client's model for 250 epochs and employed the average output digits from all client models as the ensemble strategy.
In the experiments with our method, we adapted the number of training iterations based on the volume of training data, resulting in fewer iterations for larger datasets and more for smaller ones, as in \begin{equation}
    E = 10^6 \cdot K / N,
\label{eq8}
\end{equation}
where $E$ denotes the number of DDPM training epochs, $K$ denotes the number of label categories, and $N$ denotes the number of training samples. During image generation, we used the final parameters of the DDPM as the generator, setting the output image size to $256 \times 256$. The soft label weight is adjusted based on the actual testing performance, reflecting the reliability of the training data produced by the generator.

\begin{table}
\caption{Hyperparameters for FL Algorithms}
\label{table4}
\centering
\setlength{\tabcolsep}{3pt}
\begin{tabular}{p{90pt}<{\centering}|p{115pt}<{\centering}}

\hline
\textbf{FL Algorithms} & \textbf{Corresponding Hyperparameters} \\
\hline
FedProx & $\mu = 0.01$ \\
MOON & $\mu = 1.0$ \\
FedNova & $\rho = 0.9$ \\
FedRS & $\alpha = 0.5$ \\ 
Elastic Aggregation & $\mu = 0.95, \tau = 0.5$ \\ 
PRR-FL & $\alpha_1 = 0.7, \alpha_2 = 0.9$ \\
DENSE & $\lambda_1 = 1.0, \lambda_2 = 0.5$ \\
\hline
\end{tabular}
\vspace{-0.8em}
\end{table}

\subsubsection{Metrics}
For traditional federated learning algorithms, the performance of the global model is evaluated on a pre-defined test dataset. In contrast, for personalized federated learning algorithms, the average performance of individual client models is calculated as the evaluation metric. In particular, in the context of a real-world multi-center non-IID dataset TB, the evaluation of traditional federated learning algorithms involves calculating the accuracy of the global model on the union of test sets from three regions; whereas the evaluation of personalized federated learning algorithms entails computing the mean accuracy of each client's model on its respective test set.
We utilized the top-1 accuracy on the test dataset as the primary metric for our benchmark on federated learning algorithms. All algorithms were executed for the same number of global and local epochs to ensure a fair comparison.

\section{Experimental Results and Analysis}

\subsection{Performance Comparison of Classification Models}
As shown in Table~\ref{table5} and Table~\ref{table6}, all federated learning algorithms experience varying degrees of accuracy reduction compared to centralized model training. 
FedAvg demonstrated a significant performance discrepancy compared to centralized training on the TB, Retino, and DR datasets, likely due to the insufficient data volume on individual clients. 
MOON, FedProx, and Elastic Aggregation demonstrate relatively superior and stable performance improvements over FedAvg among the traditional federated learning optimization algorithms which adjust for aggregation methods and local client training. Elastic Aggregation outperformed FedAvg across five datasets in the experiment, and achieved the optimal results on one of the datasets and second-best results on three others. Both FedProx and MOON also outperformed FedAvg across six datasets, with each achieving the best result on one dataset and second-best results on two datasets. 

FedBN showed broad advantages over the pFedAvg baseline (i.e., simply using clients' models for evluation under FedAvg algorithm) across six datasets in the experiments, though it did not attain best results on any specific dataset. 
Both the Personalized Retrogress-Resilient framework and the DENSE framework did not perform well across medical imaging datasets in our experiments. Compared to the baseline FedAvg, the model performance declined, indicating poor adaptability in medical imaging contexts in our benchmark. 
Moreover, one-shot federated learning algorithm DENSE demonstrates notable disadvantages compared to traditional methods of multi-iteration model parameter communication and aggregation.

In general, our method of using DDPM to augment datasets yields the best classification model performance among these algorithms, which is close to the performance of centralized training. Specifically, analysis of each dataset revealed that the data augmentation with DDPM led to a negative impact on model performance only in the ColonPath dataset, where the large volume of training data already yielded good results under FedAvg, and no other algorithms improved the accuracy. Conversely, in smaller datasets like Retino, Breast, and NeoJaundice, our method with training data supplement achieved the highest performance gains. Similarly, in the TB dataset, which features a smaller samples size and a real-world non-IID scenario, training data supplement by DDPM resulted in significant performance improvements. Following the use of soft labels on generated data, the final test set accuracy exhibited better performance across most datasets, achieving optimal results in multiple instances. 

Overall, no single federated learning optimization algorithm achieves the ultimate performance across all datasets; in some cases, the optimized algorithms performs worse than the baseline FedAvg, as indicated by the arrow symbols in Table~\ref{table5} and Table~\ref{table6}. By contrast, our method with data augmentation enhances the performance of federated learning more broadly. In Tables~\ref{table7}, we summarize the number of times each algorithm achieves the optimum among all methods in our experiments, along with the average of its performance metrics. In experiments conducted across nine datasets, our method yielded seven optimal results and all outperformed the baseline FedAvg. Furthermore, our method achieves optimal performance in terms of average accuracy metrics.

\renewcommand\arraystretch{1.5}%保证每列高度是原先的1.5倍
\begin{table*}%星号表示双栏
\caption{Accuracy Results of FL Algorithms on Datasets with Cross-Validation}
\label{table5}
\centering
 \begin{tabular}{p{3.9cm}<{}p{2.3cm}<{}p{2.3cm}<{}p{2.3cm}<{}}%设定每列的宽度以及对齐方式，并且可以做到自动换行
 
    \Xhline{1.2pt}%第一条粗线
   
        & \textbf{ColonPath} & \textbf{NeoJaundice} & \textbf{Retino} \\

    \Xhline{1.2pt}%第二条粗线
        Centralized & $99.659 \pm 0.088$ & $83.533 \pm 0.554$ & $86.036 \pm 1.134$ \\
    \Xhline{1.2pt}
         FedAvg & $99.340 \pm 0.208$ & $80.533 \pm 1.413$ & $76.134 \pm 1.867$ \\ 
         FedProx & $99.280 \pm 0.144 \downarrow$ & $81.255 \pm 2.125$ & $78.759 \pm 1.933$ \\
         FedNova & $99.241 \pm 0.182 \downarrow$ & $79.733 \pm 2.031 \downarrow$ & $78.520 \pm 1.746$ \\
         MOON & $99.181 \pm 0.335 \downarrow$ & $82.667 \pm 1.980$ & $79.211 \pm 1.968$ \\
         FedRS & $99.251 \pm 0.141 \downarrow$ & $78.400 \pm 1.867 \downarrow$ & $79.952 \pm 1.784$ \\
         Elastic Aggregation & $ 99.617 \pm 0.099$ & $80.267 \pm 1.953 \downarrow$ & $82.432 \pm 1.965$ \\
    \Xhline{1.2pt}
         pFedAvg & $98.893 \pm 0.822 $ & $80.182 \pm 1.879$ & $76.839 \pm 1.980$ \\  
         FedBN & $98.406 \pm 0.738 \downarrow$ & $81.013 \pm 2.014$ & $78.936 \pm 1.845$ \\
         PRR-FL & $98.106 \pm 0.704 \downarrow$ & $75.733 \pm 2.081 \downarrow$ & $71.441 \pm 1.871 \downarrow$ \\
    \Xhline{1.2pt}
        DENSE & $79.825 \pm 0.463$ & $67.892 \pm 2.133$ & $45.031 \pm 1.902$ \\
    \Xhline{1.2pt}
        Ours (Only DDPM) & $99.160 \pm 0.250 \downarrow$ & $81.867 \pm 1.828$ & $81.982 \pm 1.873$ \\
        Ours (DDPM + Label Smoothing) & $\bm{99.639 \pm 0.095}$ & $\bm{83.467 \pm 1.844}$ & $\bm{85.135 \pm 1.730}$ \\
  \Xhline{1.2pt}%第三条粗线
  \multicolumn{4}{p{12cm}}{The accuracy results (\%) of centralized training and various federated learning algorithms on the experimental dataset obtained through cross-validation (avg $\pm$ std over samples of all folds). Symbol $\downarrow$ indicates results worse than baseline FedAvg.}\\
  \end{tabular}
  \vspace{-0.8em}
\end{table*}

\renewcommand\arraystretch{1.5}%保证每列高度是原先的1.5倍
\begin{table*}%星号表示双栏
\caption{Accuracy Results of FL Algorithms on Other Datasets}
\label{table6}
\centering
 \begin{tabular}{p{3.9cm}<{}p{1.5cm}<{}p{1.5cm}<{}p{2.1cm}<{}p{1.5cm}<{}p{1.5cm}<{}p{1.8cm}<{}}%设定每列的宽度以及对齐方式，并且可以做到自动换行
 
    \Xhline{1.2pt}%第一条粗线
   
        & \textbf{DR} & \textbf{CRC} & \textbf{COVID-QU-Ex} & \textbf{Breast} & \textbf{TB} & \textbf{Imbalanced NeoJaundice} \\

    \Xhline{1.2pt}%第二条粗线
        Centralized & $82.809$ & $95.877$ & $96.711$ & $82.809$ & $90.132$ & -- \\
    \Xhline{1.2pt}
         FedAvg & $77.490$ & $92.493$ & $94.850$ & $84.076$ & $86.691$ & $80.045$\\ 
         FedProx & $82.401$ & $\bm{95.395}$ & $96.030$ & $81.892\downarrow$ & $86.971$ & $79.596\downarrow$\\
         FedNova & $81.855$ & $93.662$ & $95.601$ & $84.713$ & $84.532\downarrow$ & $81.390$\\
         MOON & $\bm{82.810}$ & $94.332$ & $95.996$ & $83.439 \downarrow$ & $84.532\downarrow$ & $80.942$\\
         FedRS & $81.037$ & $92.721$ & $95.601$ & $84.725$ & $81.295\downarrow$ & $79.372\downarrow$\\
         Elastic Aggregation & $80.900$ & $91.964\downarrow$ & $\bm{96.672}$ & $82.803\downarrow$ & $88.129$ & $79.821\downarrow$\\
    \Xhline{1.2pt}
         pFedAvg & $77.316$ & $92.763$ & $94.329$ & $83.821$ & $87.122$ & $79.405$\\
         FedBN & $77.626$ & $93.141$ & $95.730$ & $84.713$ & $81.775 \downarrow$ & $79.746$\\
         PRR-FL & $75.825 \downarrow$ & $91.088 \downarrow$ & $93.674 \downarrow$ & $79.522 \downarrow$ & $86.552 \downarrow$ & $80.261$\\
    \Xhline{1.2pt}
        DENSE & $47.969$ & $73.787$ & $50.491$ & $69.468$ & $66.019$ & $65.372$\\
    \Xhline{1.2pt}
        Ours (Only DDPM) & $78.353$ & $93.348$ & $95.710$ & $85.892$ & $86.694$ & $80.942$\\
        Ours (DDPM + Label Smoothing) & $81.763$ & $94.267$ & $\bm{96.672}$ & $\bm{86.624}$ & $\bm{88.489}$ & $\bm{81.839}$\\
  \Xhline{1.2pt}%第三条粗线
  \multicolumn{7}{p{16.5cm}}{The accuracy results (\%) of centralized training and various federated learning algorithms on the experimental dataset with pre-defined testing sets. DR, CRC, COVID-QU-Ex, Breast, and TB have test sets defined by the datasets provider. For NeoJaundice, we initially designated a portion as the test set, after which the remaining data were unevenly allocated as clients' training subsets, resulting in Imbalanced NeoJaundice dataset. Symbol $\downarrow$ indicates results worse than baseline FedAvg.}\\
  \end{tabular}
  \vspace{-0.8em}
\end{table*}

\renewcommand\arraystretch{1.5}%保证每列高度是原先的1.5倍
\begin{table}
\caption{Summary of Experiment Results of FL Algorithms}
\label{table7}
\centering
 \begin{tabular}{p{3.9cm}<{}p{1cm}<{}p{1cm}<{}p{1cm}<{}}%设定每列的宽度以及对齐方式，并且可以做到自动换行
 
    \Xhline{1.2pt}%第一条粗线
   
        & \textbf{Optimal Times} & \textbf{Below Baseline Times} & \textbf{Average Metrics} \\

    \Xhline{1.2pt}%第二条粗线
         FedAvg & 0 & -- & 85.74 \\ 
         FedProx & 1 & 3 & 86.84 \\
         FedNova & 0 & 3 & 86.58 \\
         MOON & 1 & 3 & 87.01\\
         FedRS & 0 & 4 & 85.82\\
         Elastic Aggregation & 1 & 3 & 86.96\\
    \Xhline{1.2pt}
         pFedAvg & 0 & -- & 85.63 \\
         FedBN & 0 & 2 & 85.68 \\
         PRR-FL & 0 & 8 & 83.58\\
    \Xhline{1.2pt}
        DENSE & 0 & 9 & 62.87\\
    \Xhline{1.2pt}
        Ours (Only DDPM) & 0 & 1 & 87.11\\
        Ours (DDPM + Label Smoothing) & 7 & 0 & \textbf{88.66}\\
  \Xhline{1.2pt}%第三条粗线
  \end{tabular}
\end{table}

\subsection{Impact of Non-IID Data}
\begin{figure}[!t]
\centerline{\includegraphics[width=\columnwidth]{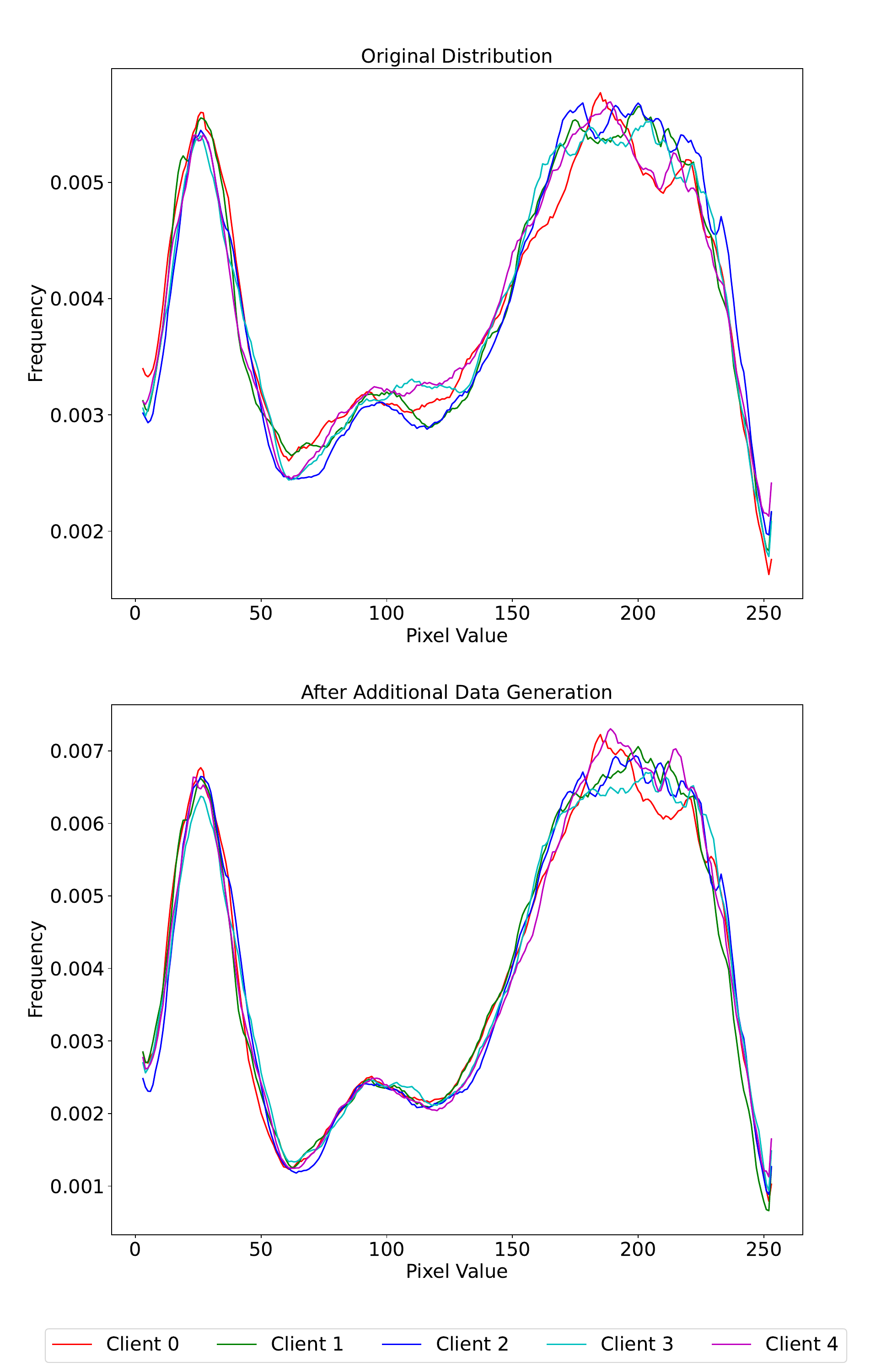}}
\caption{Distribution of image pixels in different client datasets of NeoJaundice.}
\label{fig5}
\vspace{-0.8em}
\end{figure}

To investigate the impact of using diffusion models for generating images to supplement the client training datasets on data distribution, we plotted the pixel distribution frequency curves for the datasets of different clients in each experiment. As shown in Fig.\ref{fig5}, the characteristics of medical images are marked by greater variability in pixel distribution among different images, i.e., feature shift. However, we observed that the locally augmented datasets, generated through the diffusion model, exhibit a more consistent pixel distribution across different clients.

The real-world non-IID dataset TB exhibits significant differences in pixel distribution among different client datasets. However, after augmentation using DDPM, the distributions of the local datasets from the three clients demonstrate a trend towards convergence, with a reduction in the standard deviation of the means of the pixel values across clients, as shown in Table~\ref{table8}.

\begin{figure}[!t]
\centerline{\includegraphics[width=\columnwidth]{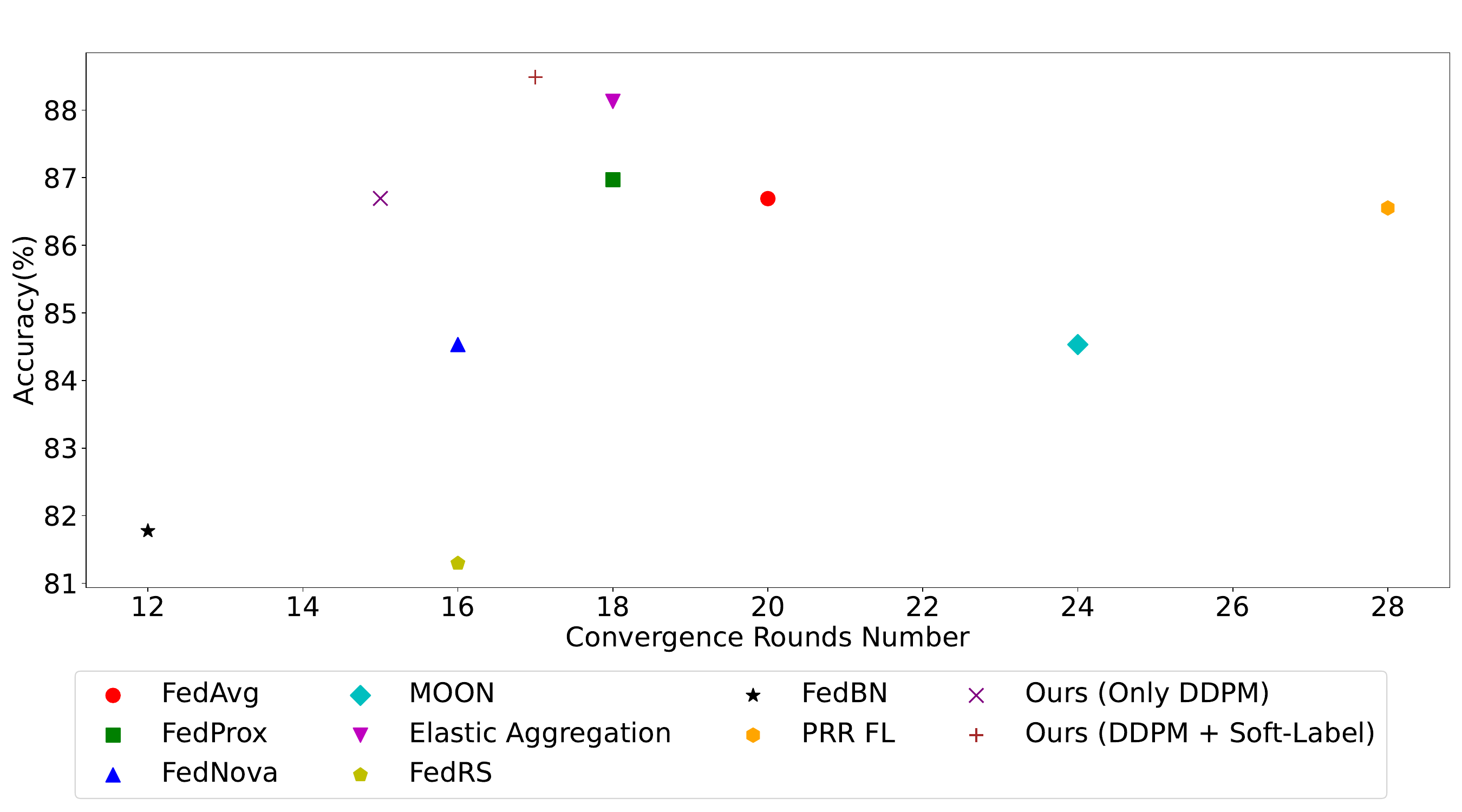}}
\caption{The trade-off between the communication cost in federated learning and the performance of the classifier. The x-axis represents the global epochs required for convergence, while the y-axis indicates the testing accuracy after convergence.}
\label{convergence}
\vspace{-0.8em}
\end{figure}

The convergence iterations and accuracy distribution on the TB dataset are illustrated in Fig.~\ref{convergence}. We observed that the non-IID training data indeed adversely affected the convergence of the global model in FedAvg, while most optimization algorithms required fewer training rounds to achieve convergence in comparison. 
In FedBN, the number of rounds required for the global model in federated learning to converge was minimized, likely due to the avoidance of negative impacts from aggregating personalized parameters associated with the batch normalization layers on the client models. However, the resulting classification model's accuracy was not satisfactory, falling below that of the baseline FedAvg. Other optimization algorithms, such as FedRS and FedNova, also achieved faster convergence rates, but the accuracy of the converged classification models decreased relative to FedAvg. Meanwhile, MOON and Personalized Retrogress-Resilient framework required more global communication rounds to reach convergence, and their final model accuracy did not surpass that of FedAvg. In contrast, FedProx and Elastic Aggregation demonstrated relatively favorable performance in both convergence speed and final accuracy of global model.
We also observed that integrating data generated by the DDPM into the training dataset alleviated the challenges of convergence for the global aggregation model caused by the scarcity and heterogeneity of client data. However, the final classification accuracy after convergence has not yet reached optimal levels, potentially due to the generated data misleading the training of client classification models to some extent. Furthermore, after incorporating label smoothing techniques to the generated data, the number of rounds required for the global model to converge increased slightly, yet it remained lower than that of FedAvg. Nevertheless, the final accuracy achieved the best results among all evaluated methods.

\begin{table}
\caption{Statistical Data of Image Pixel Values in TB}
\label{table8}
\centering
\setlength{\tabcolsep}{3pt}
\begin{tabular}{p{80pt}<{\centering}|p{80pt}<{\centering}|p{70pt}<{\centering}}
\hline
& \textbf{Mean Pixel Values of 3 Clients} & \textbf{Standard Deviation of Mean Values} \\
\hline
original & 155.1, 137.1, 110.3 & 18.4 \\ 
with generated images & 138.6, 133.4, 114.3 & 10.4 \\
\hline
\end{tabular}
\end{table}

In the scenario where the amount of local data of participants were imbalanced, simulated using the NeoJaundice dataset, the combination of diffusion generation techniques and soft label methods also yielded the most favorable results among all federated learning algorithms. We supplemented the training datasets of the three clients to an equal number of images, directly addressing the problem of training data scarcity among certain clients and the uneven distribution of client numbers, which led to aggregation bias in the model and consequently impaired the performance of the global model. The distribution of data volume for each client after the addition of generated data in the training set is shown in Fig.~\ref{subfigure6}. Overall, it appears that the impact of imbalanced data distribution on the federated learning of medical image classification models is minimal.

\subsection{FL Efficiency}
\subsubsection{Communication Cost}
In the federated learning framework, clients and servers need to share a set of parameters including model parameters in order to achieve the goal of training a robust model collaboratively while maintaining data decentralization. This exchange of data can lead to communication cost in practical federated learning scenarios.

For algorithms based on FedAvg, it is necessary to transmit all parameters of the entire model. The number of all parameters of the ResNet-50 model used in our experiments is about 24M. 
Both the Personalized Retrogress-Resilient framework and our method utilizing DDPM for dataset augmentation require only the transmission of the classification model's parameters, introducing no additional communication cost. FedBN involves relatively fewer parameters for transmission as it does not aggregate certain parameters of the batch normalization layers, eliminating the need for communication between clients and servers for this subset of parameters. In the context of the ResNet-50 model used in our experiments, the reduction in transmitted parameters compared to FedAvg is approximately 91K. In FedProx, the server must send an extra parameter to the clients at the start of federated training to control the weight of the regularization term in the loss function. 
Similarly, both MOON and FedRS require the server to transmit only one additional control parameter. Overall, there are no significant differences compared to FedAvg. In FedNova, apart from the control parameter issued by the server at the beginning of training, each client must calculate a weight value based on the number of local training epochs and a hyperparameter to transmit to the server during each global round. In Elastic Aggregation, the server needs to send a parameter at the start of training to guide the clients in calculating sensitivity updates. After local training is completed, clients must transmit a vector reflecting the sensitivity of the model parameters across various layers to the server for aggregation calculations. This incurs the highest communication cost among these optimization algorithms. Under the DENSE framework, communication between the server and clients occurs only once to transmit model parameters for integration, making it a one-shot federated learning approach that significantly reduces communication cost. 

\begin{figure}[!t]
\centerline{\includegraphics[width=1.2\columnwidth]{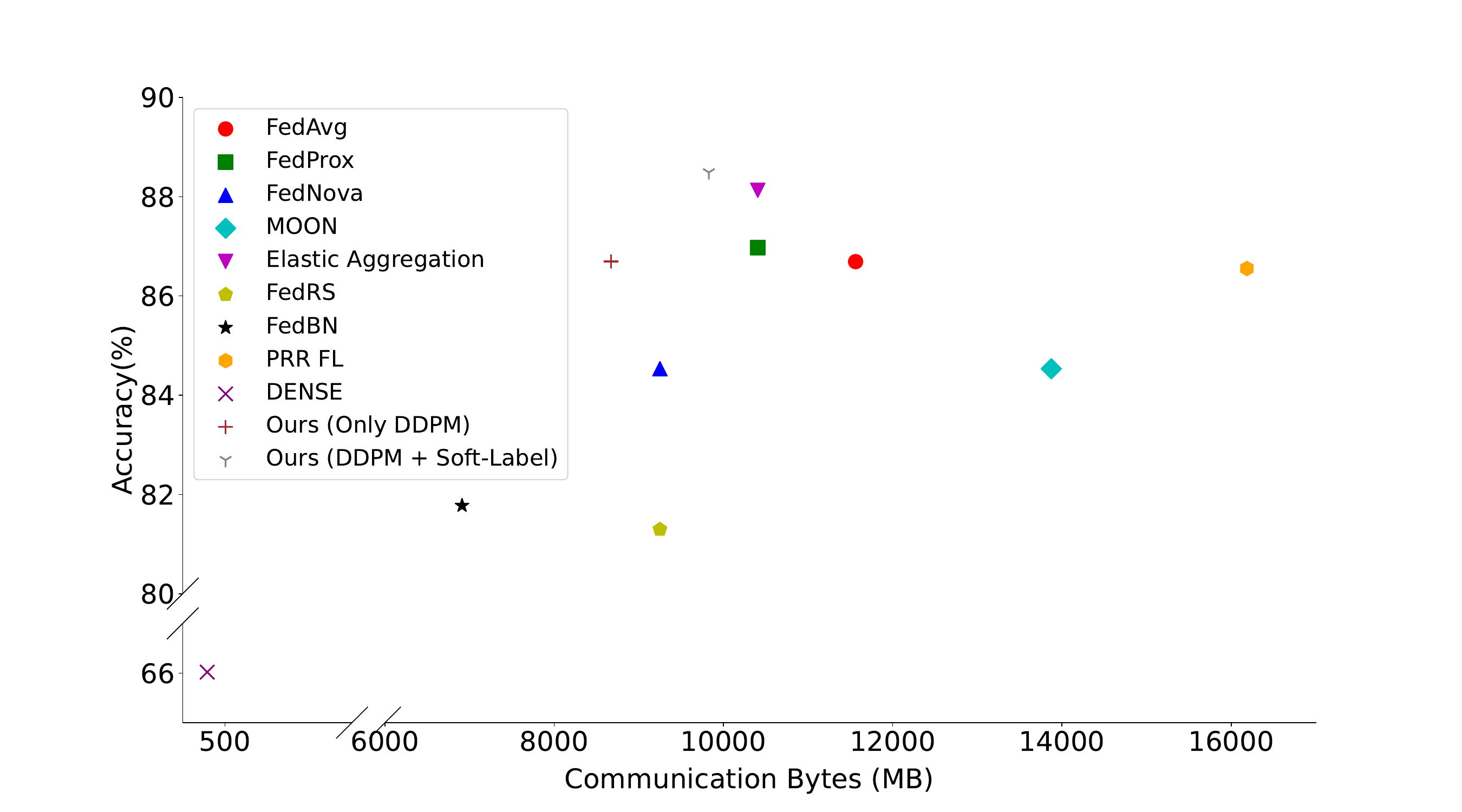}}
\caption{The trade-off between the communication cost in federated learning and the performance of the classifier. The x-axis represents
the communicate cost at convergence, measured in Mbytes, while the y-axis indicates the testing accuracy after convergence.}
\label{communication}
\vspace{-0.8em}
\end{figure}

Based on a binary classification ResNet-50 model, the detailed calculations of communication cost and the size of model in single-round communication are summarized in Table~\ref{table9}. Taking the experimental results of real-world multi-center dataset TB as an example, the relationship between accuracy after convergence and communication cost is illustrated in Fig.~\ref{communication}. While one-shot federated learning DENSE significantly reduces communication volume, its result model performance is notably inferior to that of traditional federated learning with multiple rounds of model parameters communication. Our method achieves excellent results among all federated learning algorithms in our benchmark, characterized by relatively low communication cost and high accuracy of the global model.

\renewcommand\arraystretch{1.5}%保证每列高度是原先的1.5倍
\begin{table}
\caption{Communication Cost of FL Algorithms}
\label{table9}
\centering
 \begin{tabular}{p{1.5cm}<{}p{3cm}<{}p{2.2cm}<{}}%设定每列的宽度以及对齐方式，并且可以做到自动换行
 
    \Xhline{1.2pt}%第一条粗线
   
        & \textbf{Number of Parameters in Communication} & \textbf{Size of Transmitted Model (Bytes)} \\

    \Xhline{1.2pt}%第二条粗线
         FedAvg/PRR-FL/Ours & $\num{2.409e7} \cdot E \cdot K \cdot 2 $ & $\num{9.635e7}$ \\ 
         FedProx/ MOON/FedRS & $\num{2.409e7} \cdot E \cdot K \cdot 2 + K$ & $\num{9.635e7}$  \\
         FedNova & $\num{2.409e7} \cdot E \cdot K \cdot 2 + K + E \cdot K$ & $\num{9.635e7}$  \\
         Elastic Aggregation & $\num{2.409e7} \cdot E \cdot K \cdot 2 + K + 163 \cdot E \cdot K$ & $\num{9.635e7}$  \\
         FedBN & $\num{2.399e7} \cdot E \cdot K \cdot 2$ & $\num{9.598e7}$   \\
         DENSE & $\num{2.409e7} \cdot K$ & $\num{9.635e7}$  \\
  \Xhline{1.2pt}%第三条粗线
  \multicolumn{3}{p{7.6cm}}{The number \num{2.409e7} represents the total number of parameters in the ResNet-50 model, while the number 163 indicates the count of layers with trainable parameters. $E$ denotes the number of global training epochs, and $K$ denotes the number of participating clients.}\\
  \end{tabular}
  \vspace{-1.0em}
\end{table}

\subsubsection{Computational Cost}
In the general FedAvg framework, the computational cost of federated learning primarily arises from local gradient descent training on individual clients, which includes calculating cross-entropy loss and performing backpropagation. FedBN also does not introduce any extra computational cost by contrast. 
FedProx introduces additional computations by evaluating the $L2$ distance between the global model dispatched by the server and the current local model parameters. 
MOON necessitates the storage of intermediate output representations and retaining both the local model from the end of the previous global round and the global model sent by the server. During each batch of stochastic gradient descent, it requires additional calculations to produce feature outputs from these two models based on the input samples, and subsequently computes the cosine similarity between these outputs and the current model's outputs. This introduces significant additional cost. Consequently, while MOON achieves relatively optimal experimental results, it also incurs the highest computational complexity and GPU memory usage on the client side. 
FedRS implements a scaling mechanism on the model outputs to correct the cross-entropy loss calculation, resulting in a slight increase in computational demand. 
FedNova adds extra computations by requiring clients to calculate a control parameter at the end of each global round of gradient descent training, which is then used to adjust the updated local model parameters. The server also employs this control parameter to adjust the weighted average calculations of the global model parameters during aggregation, introducing computational cost on both the client and server sides. 
Elastic Aggregation requires gradient calculations on a subset of training data before local training begins to assess the sensitivity of model parameters at various layers based on these gradient values. 
Personalized Retrogress-Resilient framework necessitates that each client maintains an additional proxy model, which must undergo gradient descent training alongside the local personalized model, resulting in increased GPU memory and computational cost. 
DENSE requires the additional training of a generator that produces synthetic data for distillation learning of the global model. 
Similarly, our method mandates that each client trains an extra DDPM to generate images. And during the iterative process of federated learning, the computational cost of our method is exactly the same as FedAvg.

In terms of FLOPs (Floating Point Operations) of federated learning models, there are major differences among MOON, Personalized Retrogress-Resilient framework, and other optimization algorithms. The statistics for FLOPs in federated learning with binary classification ResNet-50 models are presented in Table~\ref{table11}. It is indicated that our method is entirely consistent with the baseline FedAvg in terms of FLOPs.

\begin{table}
\caption{FLOPs of FL Models}
\label{table11}
\centering
\setlength{\tabcolsep}{3pt}
\begin{tabular}{p{80pt}<{\centering}|p{80pt}<{\centering}}

\hline
\textbf{Algorithms} & \textbf{FLOPs} \\
\hline
Ours & \num{3.454e11} \\ 
FedAvg & \num{3.454e11} \\ 
FedBN & \num{3.454e11} \\ 
FedNova & \num{3.454e11} \\ 
FedRS & \num{3.454e11} \\ 
FedProx & \num{3.454e11} \\ 
DENSE & \num{3.454e11} \\ 
Elastic Aggregation & \num{3.454e11} \\ 
PRR-FL & \num{6.908e11} \\
MOON & \num{1.036e12} \\

\hline
\end{tabular}
\vspace{-0.8em}
\end{table}

\section{Conclusion}
Our work addresses the lack of benchmarking for federated learning in the field of medical image classification. We experimented with a large variety of state-of-the-art federated learning methods on medical imaging datasets, showing that no single federated learning algorithm can consistently achieve optimal performance across all machine learning-based disease diagnosis scenarios. In contrast, our data augmentation method demonstrated superior federated learning outcomes across a broader range of contexts.

Based on our benchmark analysis, we propose the following choice strategies for federated learning algorithms. (1) As our method demonstrates strong applicability across most medical image classification tasks in diagnostic scenarios, it is recommended to employ our method of incorporating DDPM and label smoothing techniques for data augmentation with generated medical images, whenever computational resources and GPU capacity at the client side are sufficient. Alternatively, for traditional federated learning optimization, MOON is advised. (2) In cases where clients have limited computational and GPU memory resources that cannot support MOON or our method, a better choice would be to use Elastic Aggregation or FedProx. (3) If the network bandwidth of the federated learning system is insufficient, and there is a desire to reduce the data transfer volume between clients and the server or to pursue a more streamlined engineering implementation, FedProx is more suitable.

Regarding the limitations of our work, we utilized a fixed number of clients in our federated learning framework without exploring the impact of client numbers on federated learning algorithms in the medical image classification tasks. These experiments could be further explored in future work.

\section*{References}
\bibliography{IEEEabrv,ref}{}
\bibliographystyle{IEEEtran}

\end{document}